\documentclass[runningheads]{llncs}
\usepackage{cite}
\usepackage{graphicx}
\usepackage{multirow}
\usepackage{amssymb}
\usepackage{array}
\usepackage{url}
\usepackage{enumerate}
\usepackage{enumitem}
\usepackage{amsmath}
\usepackage{booktabs}
\usepackage{multirow}
\usepackage{filecontents}
\usepackage{hyperref}
\usepackage{marvosym}
\usepackage{caption}
\usepackage{pgffor}
\usepackage{algorithm}
\usepackage{algorithmic}
\usepackage{subcaption}
\usepackage{xcolor} 
\definecolor{darkgreen}{rgb}{0.66, 0.0, 0.0}

\begin{document}

\title{Frequency-Calibrated Membership Inference Attacks on Medical Image Diffusion Models}
\author{Xinkai Zhao$^1$\thanks{This work was completed during an internship at Preferred Networks, Inc.}
, Yuta Tokuoka$^2$, Junichiro Iwasawa$^2$, and Keita Oda$^2$}
\institute{$^1$ Graduate School of Informatics, Nagoya University, Japan\\
$^2$ Preferred Networks, Inc., Japan
}
\titlerunning{Frequency-Calibrated Membership Inference Attacks}
\authorrunning{X. Zhao et al.}

\maketitle              
 
\begin{abstract}
The increasing use of diffusion models for image generation, especially in sensitive areas like medical imaging, has raised significant privacy concerns. 
Membership Inference Attack (MIA) has emerged as a potential approach to determine if a specific image was used to train a diffusion model, thus quantifying privacy risks. 
Existing MIA methods often rely on diffusion reconstruction errors, where member images are expected to have lower reconstruction errors than non-member images.
However, applying these methods directly to medical images faces challenges. 
Reconstruction error is influenced by inherent image difficulty, and diffusion models struggle with high-frequency detail reconstruction. 
To address these issues, we propose a Frequency-Calibrated Reconstruction Error (FCRE) method for MIAs on medical image diffusion models.  
By focusing on reconstruction errors within a specific mid-frequency range and excluding both high-frequency (difficult to reconstruct) and low-frequency (less informative) regions, our frequency-selective approach mitigates the confounding factor of inherent image difficulty.
Specifically, we analyze the reverse diffusion process, obtain the mid-frequency reconstruction error, and compute the structural similarity index score between the reconstructed and original images. Membership is determined by comparing this score to a threshold. 
Experiments on several medical image datasets demonstrate that our FCRE method outperforms existing MIA methods. 
\keywords{Diffusion Models \and Privacy Preserving \and Membership Inference Attack \and Difficulty Calibration.}
\end{abstract}

\section{Introduction}

Denoising diffusion models \cite{ho2020denoising, song2020denoising}, which leverage diffusion processes to learn data distributions and generate high-quality images, have emerged as a powerful tool in medical image processing \cite{kazerouni2023diffusion}.
Given the sensitive nature of medical data and its limited accessibility, there has been a surge of interest in utilizing generative models for data sharing \cite{ayyoubzadeh2022clinical, chang2020synthetic} and creating synthetic datasets for neural network training \cite{daum2024differentially, liu2024generating, ye2023synthetic}.
However, recent studies have revealed that diffusion models used for medical image generation may inadvertently leak training data information, raising significant patient privacy concerns \cite{luo2024exploring, akbar2025beware, wang2024semantic}.
Consequently, developing methods to detect whether specific medical images were used in training image generation models has become critically important.

Membership Inference Attack (MIA) \cite{shokri2017membership}, which determines whether specific data samples were used during model training, provides a quantitative assessment of potential privacy leakage in generative models, particularly those processing sensitive patient information.
Early approaches \cite{matsumoto2023membership} to MIA against diffusion models relied on comparing loss functions to determine training set membership using Denoising Diffusion Implicit Models (DDIM) \cite{song2020denoising}.
Recent investigations \cite{mokady2023null, zhang2024exact} have revealed that the approximation of implicit functions introduces per-step errors during DDIM inversion. 
These errors accumulate progressively, resulting in discrepancies between diffusion and denoising trajectories.
Therefore, State-of-the-Art (SOTA) MIA methods have achieved superior performance by analyzing the pixel-level symmetry between diffusion and denoising trajectories in DDIM \cite{duan2023diffusion, kong2023efficient} .
Specifically, DDIM exhibits better inversion symmetry for images from the training set compared to unseen images, so it is able to better reconstruct the original image after noise addition and denoising.

While these methods have demonstrated effectiveness in performing MIA on natural images \cite{duan2023diffusion, kong2023efficient}, their direct application to medical imaging datasets, such as MRI and X-ray scans, shows limited success in distinguishing training set membership. 
This performance degradation can be attributed to the distinct frequency domain characteristics of medical images. 
Unlike natural images, which predominantly contain medium-frequency components, medical images exhibit a more polarized frequency distribution, characterized by high-frequency foreground details and extensive low-frequency background regions. 
The high-frequency components, characterized by sharp pixel-to-pixel variations, introduce significant errors when analyzed at the pixel level, particularly in the context of diffusion models \cite{zhong2023patchcraft}.
Conversely, the low-frequency background regions yield small discernible errors, collectively reducing the overall inversion asymmetry of the image. 
This highlights the importance of considering sample difficulty when performing MIAs. Simpler non-member samples can exhibit similarity to members, leading to misclassification.

\begin{figure*}[t]
	\centering
	\includegraphics[width=0.95\textwidth]{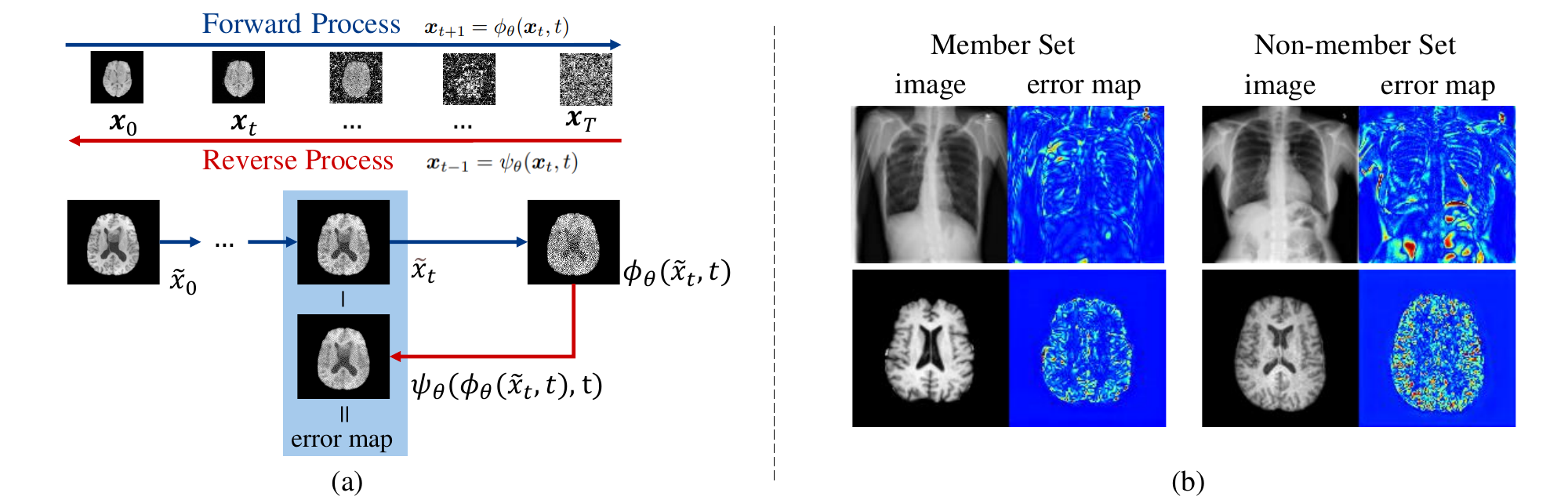}
	\caption{Background and motivation of this work. (a) illustrates the calculation of reconstruction error in the diffusion-based inversion process. (b) Images belonging to the training set (member set) generally exhibit lower reconstruction errors compared to those outside the training set (non-member set). However, accurate reconstruction of high-frequency details remains challenging for both member and non-member set images, as evidenced by the error maps.}
        \label{fig1}
\end{figure*}

The concept of difficulty calibration, as extensively studied \cite{he2024difficulty, carlini2022membership}, is critical for understanding and addressing the challenges in membership inference attacks. 
As Fig.~\ref{fig1},
it's been observed that relying solely on reconstruction error (or reconstruction similarity) is not sufficient to determine training set membership accurately. Some data samples are inherently more difficult to reconstruct due to their complex structure or uncommon features, regardless of whether they were seen during training. This means that traditional MIAs, which rely solely on reconstruction error, can be misled by these difficult non-member images.

To address this, inspired by \cite{watsonimportance, he2024difficulty, zhong2023patchcraft}, we propose a Frequency-Calibrated Reconstruction Error (FCRE) method to tackle inherent image difficulty by strategically analyzing reconstruction errors within a selected mid-frequency range, providing a more reliable measure of membership by mitigating the impact of varying difficulty levels across images. This is achieved through frequency-selective reconstruction analysis: by excluding high-frequency regions prone to reconstruction errors and excluding low-frequency regions that tend to be less informative, FCRE prioritizes the mid-frequency range to capture more discriminative features. By focusing on the mid-frequency range, the method effectively calibrates for varying levels of difficulty, reducing sensitivity to image characteristics that may confound membership inference. Furthermore, FCRE computes the SSIM score, providing a calibrated measure of reconstruction fidelity that is less sensitive to inherent image difficulty, unlike methods relying solely on pixel-wise error. Analyzing the reverse diffusion process within the mid-frequency band, FCRE captures subtle differences between member and non-member images. 
This strategic focus and the leveraging of SSIM distinguishes between members and difficult-to-reconstruct non-members, leading to enhanced attack performance, providing a more robust and generalizable method for membership inference.

In summary, our contributions are:

\begin{itemize}
    \item We address a critical need for privacy assessment in medical imaging by proposing the first MIA method specifically designed for diffusion models trained on medical images.
    \item We introduce a frequency-calibrated reconstruction error approach that leverages the spectral characteristics of medical images and mitigates the confounding effects of inherent image difficulty. Specifically, we innovatively apply difficulty calibration to the diffusion model MIA and design a novel frequency-based calibration method for medical images.
    \item Through extensive experiments on multiple medical image datasets, we demonstrate that FCRE outperforms existing methods, establishing a new benchmark for membership inference in medical image generation.
\end{itemize}

\section{Method}

In this section, we introduce our FCRE method designed for conducting MIA on medical image diffusion models. Existing reconstruction-error-based MIAs struggle with medical images due to \emph{inherent image difficulty}. FCRE addresses this by incorporating a frequency-based calibration, focusing on specific spatial frequency ranges to improve robustness.

 \subsection{Diffusion Reconstruction Error}
 
FCRE, similar to the Step-wise Error Comparing Membership Inference (SecMI) attack \cite{duan2023diffusion}, leverages the reverse process of diffusion model DDIM \cite{song2020denoising} to perform membership inference, by assessing posterior estimation with deterministic sampling and reversing.
Given an image $\mathbf{x}_0$, following standard DDIM notation, the forward process adds noise over $T$ steps:
\begin{equation}
 \mathbf{x}_t = \sqrt{\bar{\alpha}_t}\,\mathbf{x}_0 + \sqrt{1 - \bar{\alpha}_t}\;\boldsymbol{\epsilon},
\end{equation}
 where $\boldsymbol{\epsilon} \sim \mathcal{N}(\mathbf{0}, \mathbf{I})$, $\alpha_t = 1 - \beta_t$, and $\bar{\alpha}_t = \prod_{i=1}^{t} \alpha_i$. The model, parameterized by $\theta$, predicts the noise: $\hat{\boldsymbol{\epsilon}}_\theta(\mathbf{x}_t, t) \approx \boldsymbol{\epsilon}$.
 
As shown in Fig.~\ref{fig1}, FCRE uses a partial reverse process. 
Starting from an image $\mathbf{x}_0$, a multi-step DDIM denoising generates $\mathbf{x}_{t}$. Inspired by DDIM inversion analysis \cite{mokady2023null, zhang2024exact}, 
FCRE then add noises to ${\mathbf{x}}_{t+\Delta t} = \phi_\theta(\mathbf{x}_t, t)$, and reconstructs back to an approximation of the previous time step, 
\begin{equation}
\tilde{\mathbf{x}}_{t} = \psi_\theta(\phi_\theta(\mathbf{x}_t, t), t),
\end{equation}
where $\phi_\theta$ and $\psi_\theta$ are the deterministic reverse and sampling functions, respectively.

Like SecMI, FCRE relies on the hypothesis that member images exhibit better inversion symmetry when reconstructed at the same noise level. 
SecMI defines an error map $M_{error}$ to measure the approximated posterior estimation error at step t using the L2 distance (Euclidean norm): $M_{error} = \|\tilde{\mathbf{x}}_{t} - \mathbf{x}_t\|_2$

\subsection{Frequency-Based Difficulty Calibration}

Reconstruction error is affected by both membership and inherent image difficulty, a major challenge with medical images, which often exhibit polarized frequency distributions. High frequencies are harder to reconstruct, while low frequencies offer less discriminatory information.

Inspired by difficulty calibration research \cite{he2024difficulty, carlini2022membership}, FCRE uses a frequency-based calibration. We focus on the mid-frequency range, which balances reconstruction fidelity and discriminatory power, by selecting image patches within this range.
A difficulty score, representing frequency content, is calculated for each patch using the Laplacian score \cite{gonzalez2009digital}:

\begin{equation}
L(p) = \sum_{i,j} (\nabla^2 p(i,j))^2,
\end{equation}
where $p$ is a patch, $\nabla^2$ is the Laplacian, and the sum is over all pixels $(i, j)$ in the patch.  Crucially, Laplacian scores are calculated on the original image $\mathbf{x}_0$ to guide patch selection in both $\mathbf{x}_{t}$ and $\tilde{\mathbf{x}}_{t}$.

Patches are selected within the mid-frequency range based on thresholds applied to the mean absolute Laplacian score.  This excludes the area of scores above  $L_{max}$ and below $L_{min}$ in the patches. A binary mask $L(p)$ is created, identifying these mid-frequency patches, and applied to both $\mathbf{x}_{t}$ and $\tilde{\mathbf{x}}_{t}$.
\begin{equation}
\mathbf{x}_{t}^{F} = L(p) \cdot \mathbf{x}_{t},
\end{equation}
\begin{equation}
\tilde{\mathbf{x}}_{t}^{F} = L(p) \cdot \tilde{\mathbf{x}}_{t}.
\end{equation}

\subsection{SSIM-Based Membership Inference}

Following frequency-based calibration, corresponding mid-frequency patches from \(\mathbf{x}_{t}\) and \(\tilde{\mathbf{x}}_{t}\) are compared using the Structural Similarity Index (SSIM) \cite{wang2004image}, a robust perceptual metric that assesses the structural similarity between two images by considering their mean, variance, and covariance together. 
Let $p_{t}$ and $\tilde{p}_{t}$ denote corresponding patches from $\mathbf{x}_{t}^{F}$ and $\tilde{\mathbf{x}}_{t}^{F}$ respectively. The SSIM between these patches is calculated as:
\begin{equation}
SSIM(p_{t}, \tilde{p}_{t}) = \frac{(2\mu_{p_{t}}\mu_{\tilde{p}_{t}} + C_1)(2\sigma_{p_{t}\tilde{p}_{t}} + C_2)}{(\mu_{p_{t}}^2 + \mu_{\tilde{p}_{t}}^2 + C_1)(\sigma_{p_{t}}^2 + \sigma_{\tilde{p}_{t}}^2 + C_2)},
\end{equation}
where $\mu$ is the mean, $\sigma^2$ is the variance, $\sigma_{p_{t}\tilde{p}_{t}}$ is the covariance, and $C_1$, $C_2$ are stabilizing constants.

SSIM values range from -1 to 1, with 1 indicating perfect similarity. The MIA score uses (1-SSIM) to convert this into a distance measure, which is then added to the L2 loss:
\begin{equation}
\text{MIA score} = (1-\text{SSIM}) + \|\tilde{\mathbf{x}}_{t}^{F} - \mathbf{x}_t^{F}\|_2.
\end{equation}

SSIM focuses on structural similarity, making the score more perceptually relevant, and L2 loss captures pixel-level reconstruction errors, providing sensitivity to differences not captured by SSIM alone. 
The combination of both metrics can make the membership inference more robust to different types of distortions or artifacts introduced during the diffusion or reconstruction process and provide more accurate membership inference.

\section{Experiments}

\subsection{Datasets and Data Partition}

To comprehensively evaluate our proposed FCRE method, we conducted experiments on three diverse datasets: FeTS 2022 \cite{pati2021federated} (brain MRI), ChestX-ray8 \cite{wang2017chestx} (chest X-rays), and CIFAR-10 \cite{krizhevsky2009learning} (natural images). 
All partitioning ensures that the non-member sets are entirely unseen during training.

\begin{enumerate}
\item \textbf{FeTS 2022} \cite{pati2021federated}:
The FeTS 2022 dataset is a multi-institutional collection of brain MRI scans.
We utilized 740 cases from 16 institutions for the member set and 511 cases from a held-out institution for the non-member set.
We extracted 50 transverse T1 slices from each 3D MRI volume, resulting in 37,000 member images and 25,600 non-member images. 

\item \textbf{ChestX-ray8} \cite{wang2017chestx}:
For the ChestX-ray8 dataset, which consists of chest X-ray images, we independently and randomly selected over 3,600 images for the member set and another 3,600 images for the non-member set. This ensures that the two sets are distinct and mutually exclusive.

\item \textbf{CIFAR-10} \cite{krizhevsky2009learning}:
Following prior work \cite{duan2023diffusion, kong2023efficient}, we divided the CIFAR-10 dataset into member and non-member sets, each containing 25,000 images. Notably, we directly utilized the pre-defined data splits and publicly available weights provided by the previous study \cite{duan2023diffusion}. This allows for a fair comparison with existing methods. 
\end{enumerate}

\subsection{Implementation Details}

We implemented our FCRE MIA method to diffusion models.
For each diffusion model, we used DDIM \cite{song2020denoising} with $T = 1000$ diffusion steps and $k = 100$ sampling steps, consistent with prior work \cite{duan2023diffusion, kong2023efficient, mokady2023null}. All experiments were conducted on a single NVIDIA V100 GPU.
For experiments on FeTS 2022 and ChestX-ray8, we trained diffusion models using the Adam optimizer with a learning rate of $1 \times 10^{-4}$ and a batch size of 64, with a resolution of $128 \times 128$. 
When computing the Laplacian score, we performed statistics using \(8 \times 8\) patches.
Models were trained for 300,000 steps before conducting membership inference attacks (MIA). This training duration strikes a balance between achieving high-quality image generation and avoiding overfitting due to excessive training.

Performance was evaluated using standard MIA metrics \cite{shokri2017membership, duan2023diffusion, kong2023efficient}: AUC (Area Under the ROC Curve), ASR (Attack Success Rate), and TPR@FPR1\% (True Positive Rate at 1\% False Positive Rate).

\subsection{Results}

Table~\ref{tab:results} presents the results of membership inference attacks (MIA) on three datasets. We compare our FCRE method against state-of-the-art baselines (Loss-based \cite{matsumoto2023membership}, SecMI \cite{duan2023diffusion}, PIA/PIAN \cite{kong2023efficient}). Additionally, we include two variants of our method: FCRE (L2) (using L2 distance) and FCRE L2+SSIM) (our proposed method leveraging SSIM and L2 distance together).

\begin{table}[t]
\centering
\scriptsize
\caption{Results on different datasets. Bold values indicate the best performance for each dataset and metric.}
\label{tab:results}
\renewcommand{\arraystretch}{1.1}
\setlength{\tabcolsep}{4pt}
\begin{tabular}{lccc ccc ccc}
\toprule
\multirow{2}{*}{Method} & \multicolumn{3}{c}{FeTS 2022 \cite{pati2021federated}} & \multicolumn{3}{c}{ChestX-ray8 \cite{wang2017chestx}} & \multicolumn{3}{c}{CIFAR-10 \cite{krizhevsky2009learning}} \\
\cmidrule(lr){2-4} \cmidrule(lr){5-7} \cmidrule(lr){8-10}
& ASR & AUC & TPR$_{1\%}$ & ASR & AUC & TPR$_{1\%}$ & ASR & AUC & TPR$_{1\%}$ \\
\midrule
Loss-based \cite{matsumoto2023membership}  & 0.554  & 0.428  & 0.001 & 0.521  & 0.390  & 0.003 & 0.738  & 0.804  & 0.049 \\
SecMI \cite{duan2023diffusion}       & 0.734  & 0.708  & 0.052 & 0.856  & 0.907  & 0.176 & 0.774  & 0.839  & 0.085 \\
PIA \cite{kong2023efficient}         & 0.787  & 0.825  & 0.061 & 0.587  & 0.601  & 0.000 & 0.809  & 0.878  & 0.156 \\
PIAN \cite{kong2023efficient}        & 0.786  & 0.811  & 0.078 & 0.590  & 0.604  & 0.000 & 0.800  & 0.849  & \textbf{0.291} \\
\midrule
FCRE (L2) & 0.835  & 0.898  & 0.185 & 0.914  & 0.958  & 0.316 & \textbf{0.810}  & \textbf{0.880}  & 0.162 \\
FCRE (L2+SSIM) & \textbf{0.853}  & \textbf{0.926}  & \textbf{0.328} & \textbf{0.926}  & \textbf{0.971}  & \textbf{0.409} & 0.778  & 0.840  & 0.085 \\
\bottomrule
\end{tabular}
\end{table}

From Table~\ref{tab:results}, it is evident that our FCRE method consistently outperforms all baselines across the two medical image datasets, FeTS 2022 and ChestX-ray8. These datasets consist of high-resolution images (e.g., brain MRI and chest X-rays), where fine-grained details play a crucial role in distinguishing member and non-member samples. Our method leverages frequency calibration to effectively capture these subtle differences, significantly surpassing state-of-the-art (SOTA) methods on both datasets.

On CIFAR-10, which consists of low-resolution natural images (32×32 pixels), while PIAN achieves the highest TPR@FPR1\%, our method still performs competitively in terms of AUC and ASR. The lower resolution of CIFAR-10 limits the availability of fine-grained frequency information, which may explain the relatively smaller performance gap between our method and other baselines. Nevertheless, our results demonstrate the robustness of FCRE-MIA across diverse image types, from high-resolution medical images to low-resolution natural images.
Additionally, the superiority of FCRE (L2+SSIM) over FCRE (L2) further demonstrates the importance of perceptual similarity measures like SSIM in capturing meaningful image features.

\begin{figure*}[t]
\centering
\includegraphics[width=0.99\textwidth]{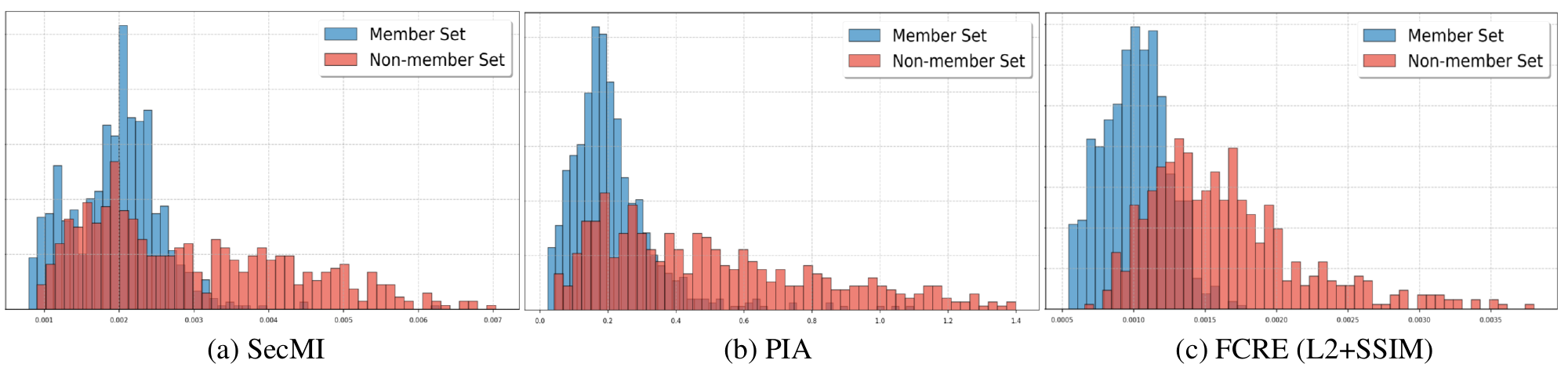}
\caption{Distributions of membership scores calculated using (a) SecMI, (b) PIA, and our proposed (c) FCRE method. The height of each bar represents the probability density. FCRE mitigates the influence of high-frequency reconstruction errors, leading to a significant reduction in the overlap between member and non-member set distributions.}
\label{fig2}
\end{figure*}

Fig.~\ref{fig2} provides a visual comparison of membership score distributions on FeTS 2022 dataset. 
Fig.~\ref{fig2}(a,b) shows the distribution using SecMI and PIA, where there is an obvious overlap between member and non-member sets. 
In contrast, Fig.~\ref{fig2}(c) demonstrates the improved separation achieved by our FCRE method, highlighting its ability to reduce ambiguity caused by high-frequency noise.

\subsection{Ablation Studies}

\begin{table}[t]
\centering
\scriptsize  
\caption{Impact of frequency thresholds ($L_{min}$ and $L_{max}$) on membership inference performance. Bold values indicate the best performance for each dataset and metric.}
\label{tab:ablation}
\renewcommand{\arraystretch}{1.1}  
\setlength{\tabcolsep}{4pt}  
\begin{tabular}{cccccccc}
\toprule
\multirow{2}{*}{$L_{min}$} & \multirow{2}{*}{$L_{max}$} & \multicolumn{3}{c}{FeTS 2022 \cite{pati2021federated}} & \multicolumn{3}{c}{ChestX-ray8 \cite{wang2017chestx}} \\
\cmidrule(lr){3-5} \cmidrule(lr){6-8}
& & ASR & AUC & TPR$_{1\%}$  & ASR & AUC & TPR$_{1\%}$  \\
\midrule
0\%      & 100\%      & 0.791  & 0.861  & 0.144  & 0.820  & 0.877  & 0.227  \\
15\%   & 85\%   & \textbf{0.853}  & \textbf{0.926}  & \textbf{0.328}  & \textbf{0.926}  & \textbf{0.971}  & \textbf{0.409}  \\
15\%   & 100\%      & 0.791  & 0.861  & 0.144  & 0.820  & 0.883  & 0.227  \\
0\%      & 85\%   & 0.818  & 0.882  & 0.254  & 0.928  & 0.968  & 0.328  \\
\bottomrule
\end{tabular}
\end{table}

To analyze the impact of frequency range selection ($L_{min}$ and $L_{max}$), we conducted an ablation study on FeTS 2022 and ChestX-ray8. The results, shown in Table~\ref{tab:ablation}, provide key insights.

When no frequency masking is applied (i.e., $L_{min} = 0\%$ and $L_{max} = 100\%$), the performance is relatively poor across all metrics. This confirms that both very low and very high frequencies contain less discriminative information and can even introduce noise that hinders MIA performance. Excluding these extremes generally improves results, with mid-frequency components showing the highest utility for distinguishing member and non-member samples. However, excluding only one extreme (e.g., removing only low or high frequencies) leads to suboptimal performance, underscoring the importance of balancing frequency masking. 

These findings demonstrate the effectiveness of frequency-based difficulty calibration and highlight the need to carefully select the frequency range. However, it is worth noting that the optimal thresholds for $L_{min}$ and $L_{max}$ vary across datasets. In this work, we did not perform an exhaustive search for the best thresholds for each dataset, as our primary goal was to validate the general utility of frequency masking. In future work, we plan to explore adaptive thresholding methods that can automatically determine the optimal frequency range for different datasets, thereby enhancing the robustness and generalizability of our approach.

\section{Conclusion}

This paper introduces Frequency-Calibrated Reconstruction Error (FCRE), a novel membership inference attack (MIA) tailored for medical image diffusion models. FCRE addresses the limitations of existing methods by incorporating frequency-based difficulty calibration, effectively mitigating the impact of inherent image difficulty—a key challenge in medical imaging. Our approach focuses on reconstruction errors within a strategically selected mid-frequency range, which captures the most discriminative information while reducing the influence of high-frequency noise and low-frequency biases.

This work underscores the privacy vulnerabilities of diffusion models, particularly in the context of medical imaging, where sensitive patient data is at risk. It also motivates further exploration into robust, privacy-preserving generative models that can balance utility and security. By shedding light on these vulnerabilities, we hope to inspire advancements in both attack methodologies and defense mechanisms, ultimately contributing to the responsible development of healthcare technologies.

\bibliographystyle{splncs04}
\bibliography{bibliography}

\end{document}